\documentclass{article}

\usepackage{placeins} 
\usepackage{booktabs}
\usepackage{subcaption} 
\usepackage{changepage}
\usepackage{setspace}
\usepackage{multicol}  
\usepackage[preprint]{neurips_2023}
\usepackage[utf8]{inputenc} 
\usepackage[T1]{fontenc}    
\usepackage{url}            
\usepackage{booktabs}       
\usepackage{amsfonts}       
\usepackage{nicefrac}       
\usepackage{microtype}      
\usepackage{colortbl}
\usepackage[table]{xcolor}
\definecolor{darkgreen}{HTML}{3cb44b}
\definecolor{darkred}{HTML}{ff7256}
\definecolor{xdcolor}{HTML}{19b5af}
\usepackage{multirow}
\usepackage[hidelinks]{hyperref}
\usepackage{graphicx}
\usepackage{xspace}
\usepackage{geometry}
\usepackage{changepage}
\usepackage{arydshln}
\usepackage{amsmath}

\newcommand{\llm}{Xmodel-2\xspace}

\title{\llm Technical Report }
\author{
Wang Qun
\hspace{0.8em}Liu Yang
\hspace{0.8em}Lin Qingquan
\hspace{0.8em}Qu Zhijiu
\hspace{0.8em}Jiang Ling \\ \\
Xiaoduo AI Lab \\
\texttt{\{wangqun,liuyangfoam,quzhijiu\}@xiaoduotech.com}
}
\date{}

\begin{document}

\maketitle

\begin{abstract}


\llm is a 1.2-billion-parameter large language model designed specifically for reasoning tasks. Its architecture enables different model scales to share a unified set of hyperparameters, allowing for extensive experimentation on smaller models and seamless transfer of optimal configurations to larger models. To maximize training efficiency and stability, \llm employs the WSD learning rate scheduler from MiniCPM. Pretrained on 1.5 trillion tokens from diverse sources, \llm achieves state-of-the-art performance in complex reasoning and agent-based tasks, while maintaining low training costs. These results highlight the potential of efficient model design and training strategies in advancing reasoning capabilities. Model checkpoints and code are publicly available on GitHub at \url{https://github.com/XiaoduoAILab/Xmodel-2}.



\begin{figure}[ht]
\centering
\includegraphics[width=0.8\linewidth]{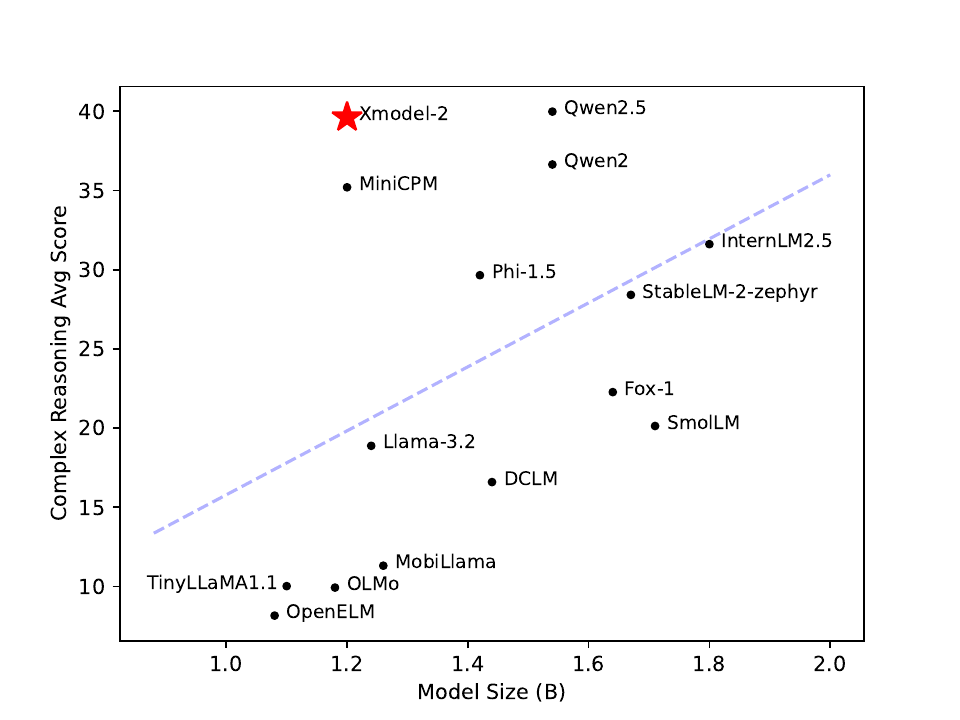}
\caption{Average Scores on Complex Reasoning Benchmarks (GSM8K, MATH, BBH, MMLU, HumanEval and MBPP).}
\label{fig:complex_arena}
\end{figure}

\end{abstract}

\section{Introduction}

Large language models (LLMs) have made significant strides in natural language understanding, demonstrating impressive performance across a range of tasks. However, they still face challenges when tackling complex reasoning tasks. Effective reasoning is crucial for applications such as automated customer service and scientific discovery \citep{ai4science2023}. While larger models typically show improved reasoning capabilities, they also require more computational resources, longer training times, and higher energy consumption.

\llm is a 1.2B-parameter model designed to balance reasoning power and training efficiency. It excels in complex reasoning, code generation, and agent-based interactions. Unlike other models, \llm incorporates an innovative architecture based on Tensor Programs \citep{yang2022tpv} \citep{yang2023tpvi} , enabling models of different scales to share the same set of hyperparameters. This approach allows for extensive hyperparameter search on smaller models, with the best configurations transferred seamlessly to larger models, enhancing both efficiency and performance.

To accelerate training and ensure stable convergence, \llm uses the Warmup-Stable-Decay (WSD) learning rate scheduler from MiniCPM \citep{hu2024minicpm}. Pretrained on 1.5 trillion tokens, \llm is capable of processing diverse inputs, such as text and code, which strengthens its performance in complex reasoning tasks.


Our contributions are as follows:
\begin{enumerate}
    \item Our \llm is open-source, aimed at improving accessibility for researchers in language model research. We believe its state-of-the-art performance and compact size make it an ideal platform for both researchers and practitioners.
    \item We advanced the decay phase by applying the WSD learning rate scheduler and exploring data ratio search, addressing a gap in the literature and achieving significant improvements in reasoning performance.
    \item We conducted a focused evaluation of \llm’s agent capabilities, demonstrating its strong potential for real-world applications such as customer service and task automation. 
\end{enumerate}

\section{Pretraining}
This chapter provides a detailed overview of the pretraining process for \llm. We begin with a description of the model architecture, followed by an explanation of the data distribution across the stable training and decay stages, and conclude with an outline of the overall training procedure.

\subsection{Model Architecture}

We adopt an architecture similar to LLama 2 \citep{touvron2023llama}, with the following configuration:

\begin{table}[ht]
  \centering
  \setlength{\tabcolsep}{6pt}
  \begin{tabular}{@{}cccccc@{}}
    \toprule
    Hidden size & Intermediate size & Attention heads &  KV heads & Layers & Context Len\\
    \midrule
    1536 & 3840 & 24 & 8 & 48 & 4096\\
    \bottomrule
  \end{tabular}
  \newline
  \caption{Model configuration for \llm.}
  \label{tab: LLM_setting}
\end{table}

\noindent\textbf{Tokenizer:} Unlike most large models that use the BPE tokenizer, \llm employs a custom Unigram tokenizer \citep{kudo2018unigram} with a vocabulary size of 65,280 tokens.

\noindent\textbf{Embedding Sharing:} In small language models (SLMs), the embedding layer constitutes a significant portion of the total parameters. To improve efficiency, we implement embedding sharing, which reduces the parameter count by 0.1B.

\noindent\textbf{Deep-and-Thin Architecture:} The importance of a deep and thin architecture for SLMs is emphasized by \citep{liu2024mobilellm}, a concept that aligns with our observations.

\noindent\textbf{Grouped-Query Attention:} To optimize training and inference efficiency, we adopt Grouped-Query Attention (GQA) \citep{ainslie2023gqa}, which utilizes 24 attention heads and 8 key-value (KV) heads.

\subsection{Training Stages}

The training of the \llm base model consists of two key stages: the Stable Training Stage and the Decay Stage.

\noindent\textbf{Stable Training Stage:} In this phase, we train on approximately 1.5 trillion tokens (see Figure~\ref{fig:datamixture} for data distribution), primarily sourced from open datasets. The training follows the optimal configuration identified through model tuning experiments, using the WSD LRS \citep{hu2024minicpm}, with a batch size of 3.93 million tokens and a maximum learning rate of 0.01.

\noindent\textbf{Decay Stage:} This stage combines the pretraining data with high-quality supervised fine-tuning (SFT) data. We apply exponential annealing to the WSD learning rate scheduler, following the formula $f(s-T) = 0.5^{(s-S)/T}$, where $T$ is set to 5000 steps (20 billion tokens), allowing the learning rate to gradually decrease during the final training phase.

\subsection{Data Ratio Optimization in the Decay Stage}
Previous work \citep{ye2024datamixinglawsoptimizing} demonstrated that data ratio experiments on small models can effectively transfer to larger models. However, their focus was mainly on cosine learning rate decay and pretraining data ratio optimization. MiniCPM \citep{hu2024minicpm} emphasized the benefits of incorporating SFT data during the decay stage, but lacked detailed exploration of data ratios under WSD learning rate schedulers.

In contrast, our work explores the interaction between SFT data and domain-specific pretraining data during the WSD decay phase. We framed the data ratio search around two key questions: the overall proportion of SFT data and the distribution of categories within SFT. This approach significantly reduces the search space, enabling efficient optimization with fewer trials.

Through over 400 trials, we identified that the optimal SFT data ratio falls between 60\% and 69\%, with the precise value depending on the internal composition of the SFT-mixed dataset. We also observed that Chain-of-Thought datasets may enhance logical reasoning \cite{suzgun2022challengingbigbenchtaskschainofthought}, while instruction-formatted datasets in mathematics and code outperform pretraining-format data in complex reasoning tasks.

\subsection{Training Data Distribution}

\begin{figure}[ht]
\centering
\begin{minipage}[t]{0.49\linewidth}
    \raggedright
    \includegraphics[width=\linewidth]{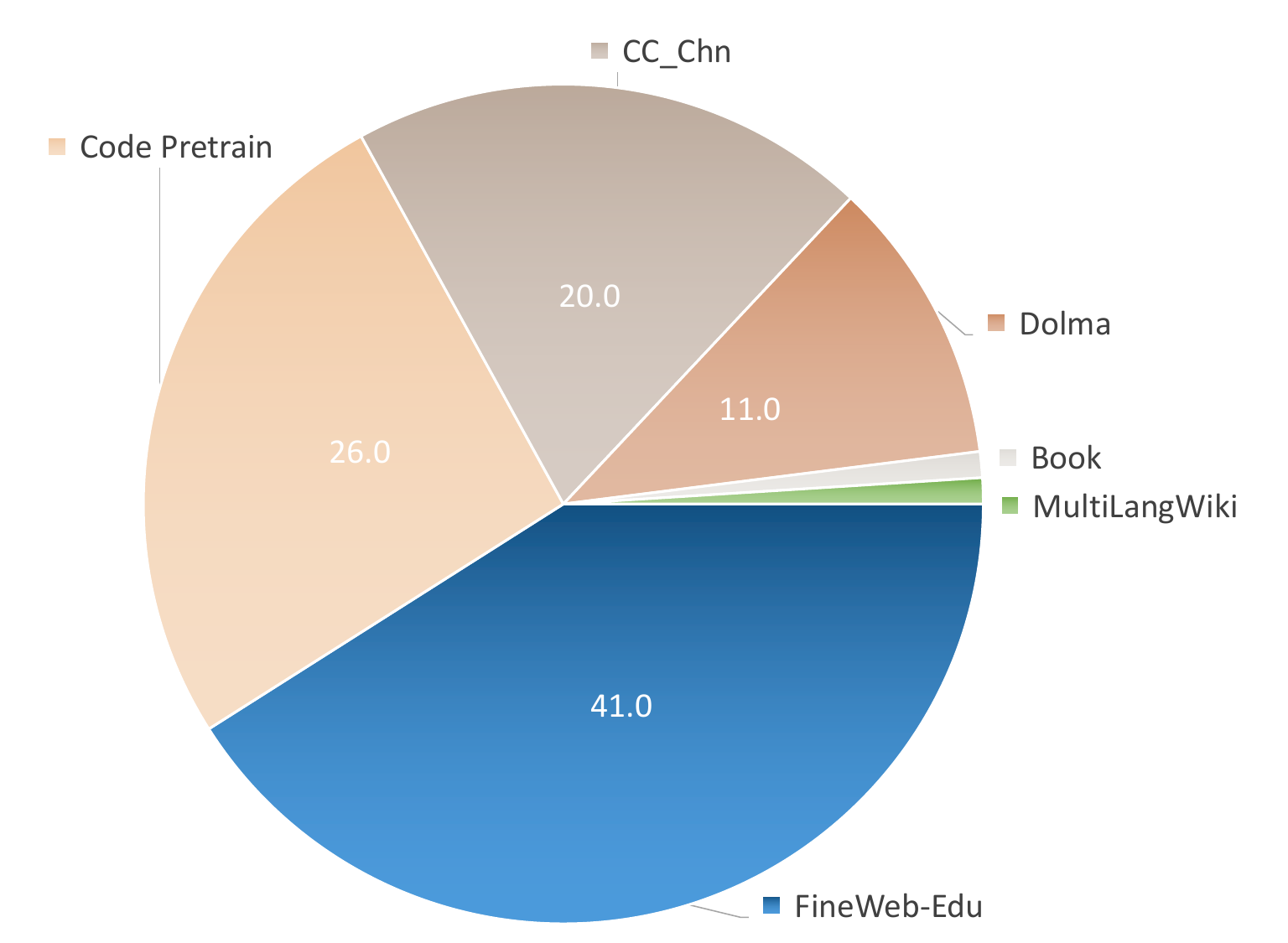}
\end{minipage}%
\hfill  
\begin{minipage}[t]{0.49\linewidth}
    \raggedright
    \includegraphics[width=\linewidth]{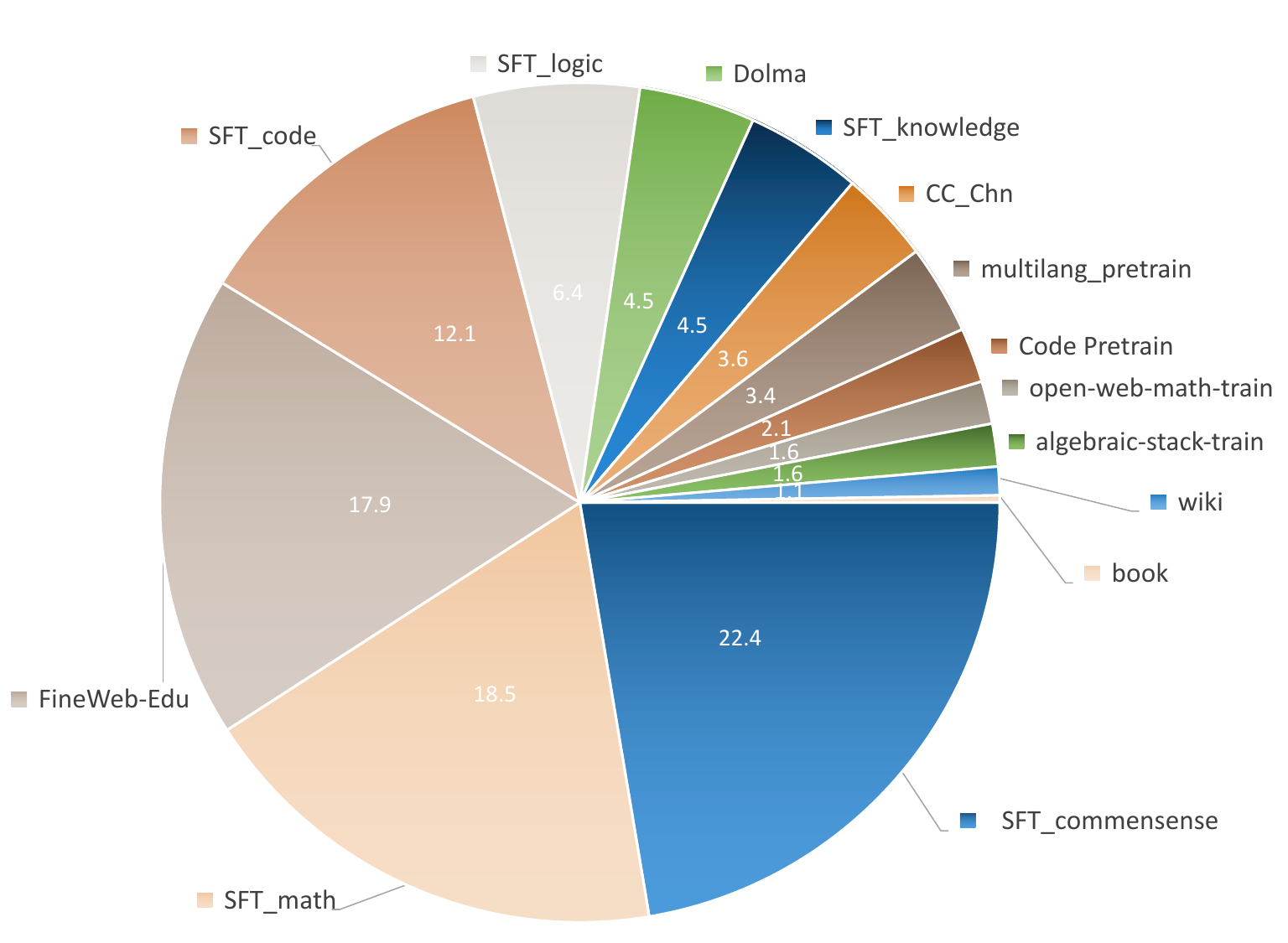}
\end{minipage}
\caption{Data mixture of different training stages.The left side represents the stable training phase, and the right side represents the decay phase.}
\label{fig:datamixture}
\end{figure}

Figure~\ref{fig:datamixture} shows the data distribution across training stages, including CC\_Chn (Chinese corpus), FineWeb-Edu \citep{penedo2024fineweb}, Dolma \citep{soldaini2024dolma} (English corpora), and Code Pretrain datasets like StarCoder \citep{li2023starcoder} and The Stack \citep{kocetkov2022stack}. The decay stage incorporates diverse data, such as EvolInstruct \citep{xu2023EvolInstruct}, OssInstruct \citep{wei2024OssInstruct}, and UltraChat \citep{ding2023UltraChat}.

The SFT-Mixed Dataset is composed of five distinct categories: Mathematics, Code, Logic, Knowledge, and Commonsense. Chain-of-Thought (CoT) data is categorized under Logic. To improve generalization, SFT prompts were diversified via rule-based transformations, though multilingual alignment and domain-specific data were excluded for future exploration. The SFT data underwent multiple rounds of SimHash deduplication with a bucket size of 1 million, improving performance by 1.7\% compared to non-deduplicated data.

Data ratio experiments revealed the effectiveness of instruction-formatted SFT data during the annealing phase, leading us to allocate 64\% to SFT data. These adjustments, combined with optimized data mixing and processing, improved complex reasoning performance by 29.31\% compared to our baseline.

\subsection{Training Loss}

\begin{figure}[ht]
\centering
\includegraphics[width=0.8\linewidth]{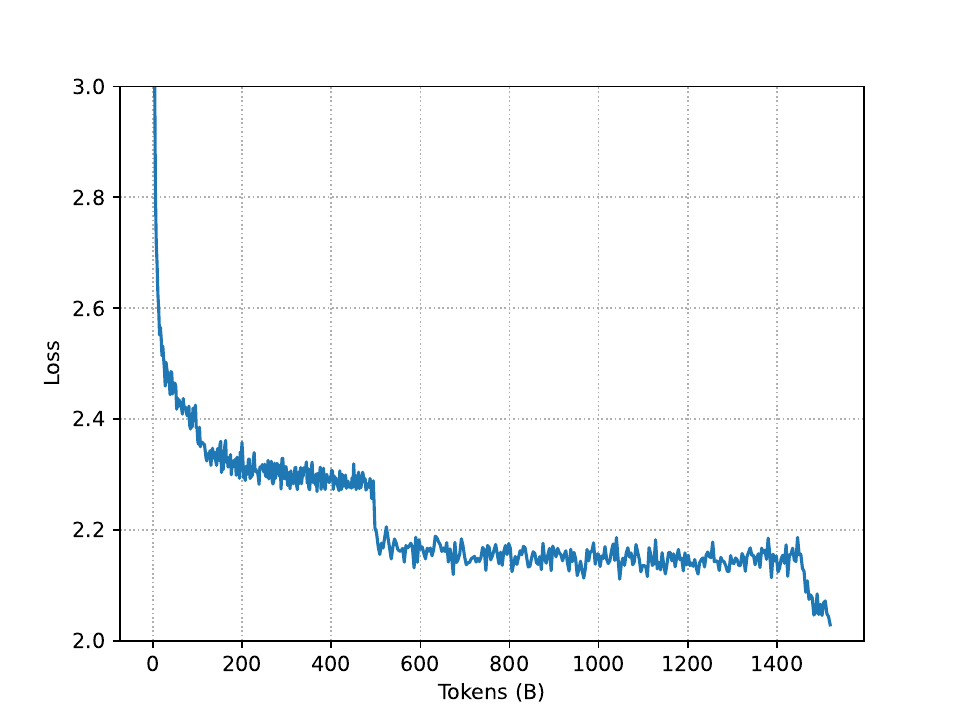}
\caption{Loss curve for \llm-1.2B.}
\label{fig:loss}
\end{figure}

Figure~\ref{fig:loss} presents the training loss curve on the FineWeb-Edu dataset \citep{penedo2024fineweb}. The initial drop corresponds to increasing the batch size from 2M to 4M, which likely replicates the stabilizing effect of a reduced learning rate \citep{smith2018batchsize}. The second drop reflects the impact of the learning rate decay phase.

\section{Results}
\label{sec:results}

We compared \llm with recent decoder-only models containing 1–2 billion parameters, as identified in \citep{lu2024slm_survey}. The baselines include TinyLLaMA \citep{zhang2024tinyllama}, InternLM2 \citep{cai2024internlm2}, Qwen2 \citep{yang2024qwen2}, MiniCPM \citep{hu2024minicpm}, Llama 3 \citep{grattafiori2024llama3}, Phi-1.5 \citep{li2023textbooks}, StableLM-2 \citep{bellagente2024stable}, OLMo \citep{groeneveld2024olmo}, MobiLlama \citep{thawakar2024mobillama}, and SmolLM \citep{allal2024SmolLM}.Our experiments demonstrate that \llm achieves \textbf{state-of-the-art (SOTA)} performance among 1B-parameter models, demonstrating the effectiveness of our training strategies and optimized data ratios, especially in commonsense reasoning, complex reasoning and agent-based tasks.

\subsection{Commonsense Reasoning}
We evaluate \llm on various commonsense reasoning benchmarks using the Language Model Evaluation Harness \citep{eval-harness}, which includes: ARC-Challenge \citep{clark2018think}, ARC-Easy \citep{clark2018think}, BoolQ \citep{clark2019boolq}, HellaSwag \citep{zellers2019hellaswag}, OpenBookQA \citep{mihaylov2018suit}, PiQA \citep{Bisk2019PIQARA}, SciQ \citep{welbl2017crowdsourcing}, TriviaQA \citep{joshi2017triviaqa}, and Winogrande \citep{Sakaguchi2021WinoGrande}. For fairness and reproducibility, all models were evaluated in the same environment using raw accuracy metrics.

Table~\ref{tab:commonsense} presents the zero-shot evaluation results, models in green perform worse than \llm, while models in red outperform \llm. In this evaluation, \llm outperforms the majority of baseline models and demonstrates competitive performance.

\begin{table}[ht]
  \centering
  \setlength{\tabcolsep}{3pt}
  \begin{tabular}{lcccccccccc}
    \toprule
    Model & \textbf{ARC-c} & \textbf{ARC-e} & \textbf{Boolq} & \textbf{HS.} & \textbf{OB.} & \textbf{PiQA} & \textbf{SciQ} & \textbf{Wino.} & \textbf{Avg}\\
    \midrule
    \rowcolor{darkgreen!20}
    MobiLlama-1B\xspace&28.24&61.53&60.92&46.74&21.80&75.14&88.20&59.27&55.23  \\
    \rowcolor{darkgreen!20}
    TinyLLaMA1.1-1.1B\xspace&30.97&61.66&55.99&46.70&25.20&72.63&89.30&59.43&55.24  \\
    \rowcolor{darkgreen!20}
    OLMo-1B\xspace&28.67&63.34&61.74&46.97&25.00&75.03&87.00&59.98&55.97\\
    \rowcolor{darkgreen!20}
    OpenELM-1.1B\xspace&28.84&62.37&63.58&48.36&25.40&74.76&90.60&61.72&56.95 \\
    \rowcolor{darkgreen!20}
    Llama-3.2-1B\xspace&31.31&65.36&63.73&47.78&26.40&74.48&91.50&61.01&57.70 \\

    \rowcolor{darkgreen!20}
    MiniCPM-1.2B\xspace&36.86&70.29&67.92&49.91&23.60&74.43&91.80&60.77&59.45 \\
    \rowcolor{darkgreen!20}
    Fox-1-1.6B\xspace&34.73&69.91&71.77&46.33&24.60&75.24&93.20&60.77&59.57\\ 
    \rowcolor{darkgreen!20}
    InternLM2.5-1.8B\xspace&35.24&66.37&79.82&46.99&22.00&73.29&94.90&62.67&60.16 \\
    \rowcolor{darkgreen!20}
    Qwen2-1.5B\xspace&33.11&66.41&72.60&48.57&27.00&75.57&94.60&65.75&60.45 \\
    \rowcolor{darkgreen!20}
    StableLM-2-zephyr-1.6B\xspace&36.52&66.79&80.00&53.26&26.80&74.86&88.00&64.09&61.29 \\
    \rowcolor{darkred!20}
    SmolLM-1.7B\xspace&43.43&76.47&65.93&49.58&30.00&75.79&93.20&60.93&61.92 \\    
    \rowcolor{darkred!20}
    Qwen2.5-1.5B\xspace&41.21&75.21&72.97&50.15&31.80&75.90&94.30&63.61&63.14 \\
     \rowcolor{darkred!20}
    DCLM-1B\xspace&41.30&74.79&71.41&53.59&32.20&76.93&94.00&66.22&63.81  \\
    \rowcolor{darkred!20}
    Phi-1.5-1.3B\xspace&44.80&76.22&74.95&47.96&38.60&76.66&93.30&72.93&65.68 \\
    
    \midrule
    \rowcolor{xdcolor!20}
    \llm-1.2B\xspace&39.16&71.55&74.65&47.45&29.20&74.81&93.60&63.93&61.79 \\
    \bottomrule
  \end{tabular}
  \newline
  \caption{Zero-shot performance on Commonsense Reasoning tasks. }
  \label{tab:commonsense}
\end{table}

\subsection{Complex Reasoning}
To evaluate the complex reasoning abilities of \llm, we conducted tests using several well-established benchmarks, including GSM8K \citep{cobbe2021gsm8k}, MATH \citep{hendrycks2021math}, BBH \citep{Suzgun2022ChallengingBT}, MMLU \citep{hendryckstest2021}, HumanEval \citep{chen2021humaneval}, and MBPP \citep{austin2021mbpp}. The first four benchmarks were assessed using the Language Model Evaluation Harness \citep{eval-harness}, while the last two were evaluated with the Code Generation LM Evaluation Harness \citep{bigcode-evaluation-harness}. The results are presented in Table~\ref{tab:complex-reasoning}.

\begin{table}[ht]
  \centering
  \setlength{\tabcolsep}{3pt}
  \begin{tabular}{lccccccc}
    \toprule
    Model & \textbf{GSM8K} & \textbf{MATH} & \textbf{BBH} & \textbf{MMLU} & 
\textbf{HumanEval} & \textbf{MBPP} & \textbf{Avg} \\
    & 5-shot& 4-shot& 3-shot& 0-shot& pass@1& pass@1 \\
    \midrule
    \rowcolor{darkgreen!20}
    OpenELM-1.1B\xspace&0.45&1.06&6.62&25.52&8.54&6.80&8.16 \\ 
    \rowcolor{darkgreen!20}
    OLMo-1B\xspace   &2.35&1.46&25.60&24.46&5.49&0.20&9.93 \\
    \rowcolor{darkgreen!20}
    TinyLLaMA1.1-1.1B\xspace   &2.50&1.48&25.57&25.35&1.83&3.40&10.02 \\
    \rowcolor{darkgreen!20}
    MobiLlama-1B\xspace   &1.97&1.54&25.76&25.26&7.93&5.40&11.31 \\
    \rowcolor{darkgreen!20}
    DCLM-1B\xspace&4.93&2.14&30.70&46.43&8.54&6.80&16.59\\ 
    \rowcolor{darkgreen!20}
    Llama-3.2-1B\xspace  &6.60&1.78&31.44&36.63&14.63&22.20&18.88\\
    \rowcolor{darkgreen!20}
    SmolLM-1.7B\xspace  &7.51&3.18&29.21&27.73&21.34&31.80&20.13\\
    \rowcolor{darkgreen!20}
    Fox-1-1.6B\xspace  &34.34&7.94&28.75&39.55&14.02&9.00&22.27\\ 
    \rowcolor{darkgreen!20}
    StableLM-2-zephyr-1.6B\xspace  &41.32&10.12&32.71&41.30&25.61&19.40&28.41\\
    \rowcolor{darkgreen!20}
    Phi-1.5-1.3B\xspace  &32.15&3.18&28.81&41.75&36.59&35.40&29.65\\
    \rowcolor{darkgreen!20}
    InternLM2.5-1.8B\xspace  &27.90&16.68&41.76&46.30&27.40&29.60&31.61 \\
    \rowcolor{darkgreen!20}
    MiniCPM-1.2B\xspace  &40.11&10.98&35.42&43.99&43.90&36.80&35.20 \\
    \rowcolor{darkgreen!20}
    Qwen2-1.5B\xspace &57.62&22.90&33.05&55.11&20.73&30.40&36.64 \\
    \rowcolor{darkred!20}
    Qwen2.5-1.5B\xspace &62.40&28.28&43.99&59.72&5.49&40.00&39.98 \\
    \midrule
    \rowcolor{xdcolor!20}
    \llm-1.2B\xspace    &55.88&25.50&48.40&48.87&29.88&29.20&39.62 \\
    \bottomrule
  \end{tabular}
  \newline
  \caption{Performance on Complex Reasoning tasks.}
  \label{tab:complex-reasoning}
\end{table}

 \subsection{Agent Capabilities}
We evaluate \llm's performance on four agent tasks using the ReAct prompting technique \citep{yao2023react}. These tasks include HotpotQA \citep{yang2018hotpotqa}, FEVER \citep{thorne2018fever}, AlfWorld \citep{shridhar2021alfworld}, and WebShop \citep{yao2023webshop}. We use EM(Exact Match) as the evaluation metric in FEVER and HotpotQA, and success rate in AlfWorld and WebShop.

To accomplish FEVER\citep{thorne2018fever} and HotpotQA\citep{yang2018hotpotqa}, the agent retrieves information from Wikipedia. In FEVER, the agent verifies the truth of a claim via multiple-choice questions, while in HotpotQA\citep{yang2018hotpotqa}, the agent reasons across multiple documents to answer complex, open-ended questions.
For AlfWorld\citep{shridhar2021alfworld}, the agent interacts with an environment of 25 containers, performing actions like retrieving or manipulating objects. This task requires spatial reasoning and decision-making.
Finally, in WebShop\citep{yao2023webshop}, the agent navigates a simulated e-commerce environment to search, customize, and purchase items. This tests the agent's ability to search efficiently and make decisions within real-world e-commerce constraints. These tasks pose significant challenges for small language models (SLMs) due to their requirements for complex reasoning, multi-step decision-making, and real-world interaction. The results are summarized in Table~\ref{tab:agent-capabilities}.

\begin{table}[ht]
  \centering
  \setlength{\tabcolsep}{3pt}
  \begin{tabular}{lccccccc}
    \toprule
    Model & \textbf{HotpotQA} & \textbf{FEVER} & \textbf{AlfWorld} & \textbf{WebShop} & \textbf{Avg} \\
    & EM& EM& success rate & success rate \\
    \midrule
    \rowcolor{darkgreen!20}
    OLMo-1B\xspace   
    &2.67&18.58&0.00&0.00&4.32 \\
    \rowcolor{darkgreen!20}
    Phi-1.5 1.3B\xspace   
    &3.54&17.56&2.24&0.80&6.04 \\
    \rowcolor{darkgreen!20}
    DCLM-1B\xspace   
    &4.92&24.39&0.75&0.00&7.52 \\
    \rowcolor{darkgreen!20}
    MobiLlama-1B\xspace   
    &0.00&30.43&0.00&0.00&7.61 \\
    \rowcolor{darkgreen!20}
    TinyLLaMA1.1-1.1B\xspace   &2.11&28.77&0.00&0.20&7.77 \\
    \rowcolor{darkgreen!20}
    OpenELM-1-1B\xspace   
    &2.70&28.37&0.00&0.40&7.87 \\
    \rowcolor{darkgreen!20}
    StableLM-2-zephyr-1.6B\xspace  &1.44&20.81&8.96&2.20&8.35\\
    \rowcolor{darkgreen!20}
    SmolLM-1.7B\xspace  
    &2.28&31.31&0.00&0.60&8.55\\
    \rowcolor{darkgreen!20}
    Fox-1-1.6B\xspace   
    &5.37&30.88&0.00&0.60&9.21 \\
    \rowcolor{darkgreen!20}
    Llama-3.2-1B\xspace  &4.87&27.67&8.21&3.20&10.99 \\
    \rowcolor{darkgreen!20}
    Qwen2.5-1.5B\xspace &13.53&27.58&5.97&0.60&11.92 \\
    \rowcolor{darkgreen!20}
    MiniCPM-1.2B\xspace  &11.00&36.57&1.60&1.00&12.52 \\
    \rowcolor{darkgreen!20}
    InternLM2.5-1.8B\xspace  &12.84&34.02&2.99&1.00&12.71 \\
    \midrule
    \rowcolor{xdcolor!20}
    \llm-1.2B \xspace    &13.70&40.00&0.78&2.20&14.21 \\
    \bottomrule
  \end{tabular}
  \newline
  \caption{Performance on Agent tasks. }
  \label{tab:agent-capabilities}
\end{table}

\section{Case Study}
\label{sec:case study}

\subsection{Calibration}

\begin{figure}[ht]
\centering
\includegraphics[width=\linewidth]{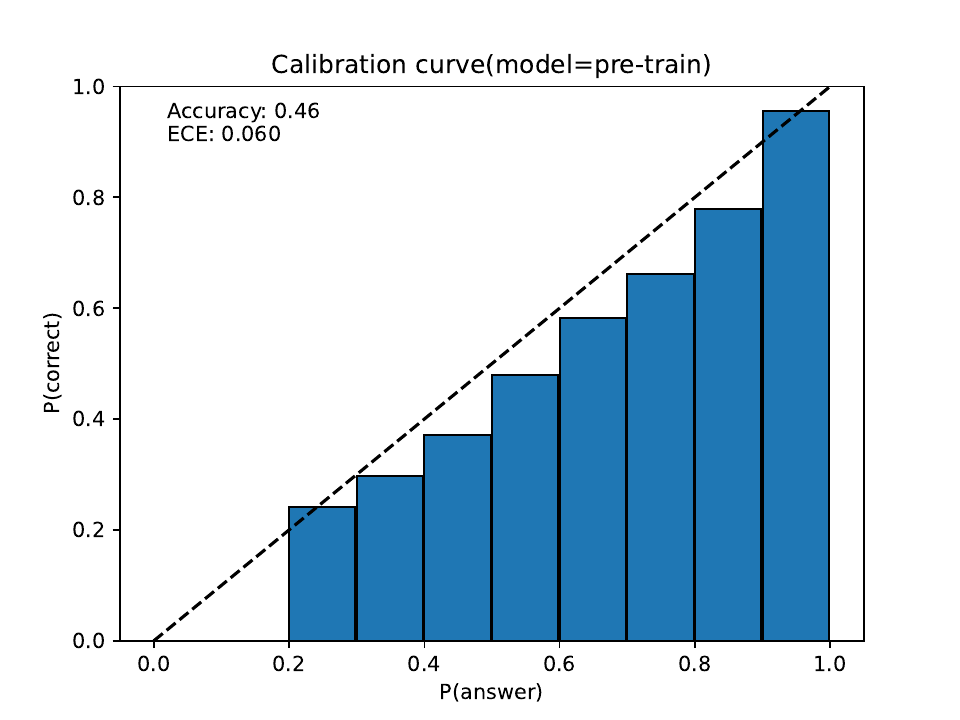}
\caption{Calibration plot for the pre-trained \llm-1.2B model on the MMLU dataset.}
\label{fig:calib}
\end{figure}

The pre-trained \llm-1.2B model exhibits strong calibration, with predicted confidence aligning closely to actual correctness probabilities. Figure~\ref{fig:calib} illustrates the calibration plot, where the x-axis represents confidence bins (log-probabilities) for A/B/C/D choices, and the y-axis shows accuracy within each bin. The dotted diagonal indicates perfect calibration.

\subsection{Post-training Scaling Law}
\begin{figure}[ht]
\centering
\includegraphics[width=\linewidth]{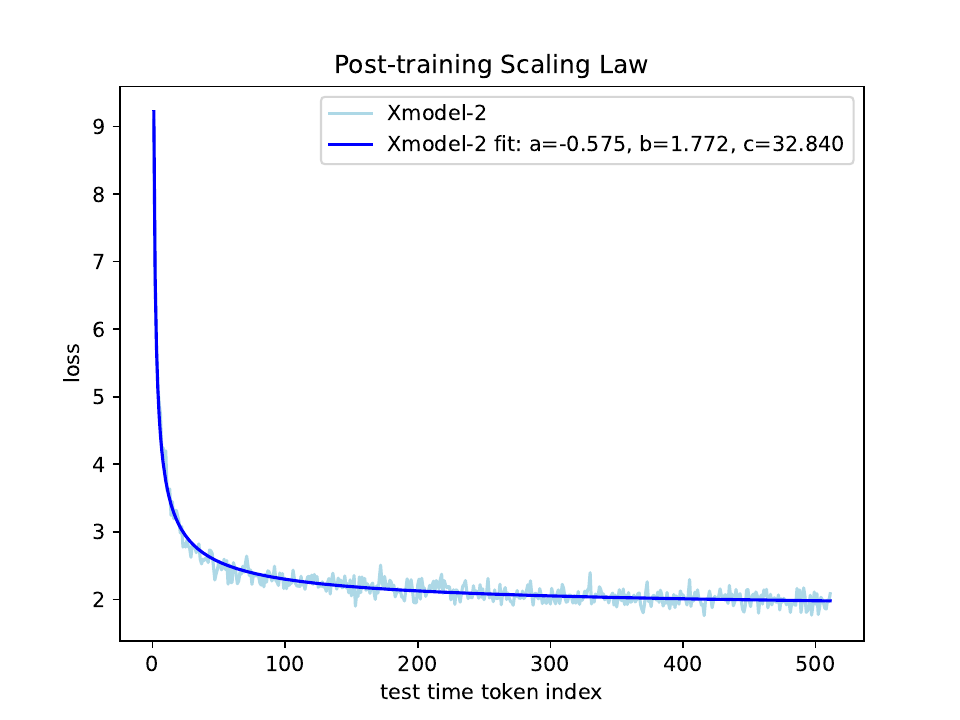}
\caption{Post-training Scaling Law for \llm-1.2B on the Wikitext-2 dataset.}
\label{fig:icl_curve}
\end{figure}

We explored the Post-training Scaling Law of \llm on the Wikitext-2 dataset, focusing on how test-time loss changes as the number of prompt tokens increases. This analysis reveals that as the context token count grows, the model’s prediction accuracy for the next token improves, with loss and token index following a power-law relationship. Figure~\ref{fig:icl_curve} shows a consistent decrease in perplexity, with diminishing returns captured by the fitted curve:

\[
L(t) = b+\left(t/c\right)^a;~~ a \sim -0.575, b\sim1.772, c\sim 32.840
\]

This suggests that, akin to OpenAI's approach of using test-time to enhance model performance, increasing context length leads to more accurate token predictions, as described by the Post-training Scaling Law.

\section{Conclusions}

This paper introduced \llm, a 1.2-billion-parameter model optimized for reasoning tasks. By leveraging the maximal update parametrization (µP), Warmup-Stable-Decay (WSD) learning rate scheduler, and data ratio optimization during the decay phase, \llm showed significant improvements in complex reasoning capabilities. Most notably, \llm achieved state-of-the-art performance in agent-based evaluations within the 1-2B parameter range, highlighting its strong potential for real-world applications such as e-commerce customer service and task automation.



\begin{spacing}{0.1} 
\footnotesize 
\begin{multicols}{2} 
\bibliography{xmodel-2}
\end{multicols}
\end{spacing}

\section{Appendix : Model Wind Tunnel Experiments}
\label{Appendix:Model Wind Tunnel Experiments}

Before pre-training, we conducted wind tunnel experiments on two small models: nano (6M) and tiny (54M) to validate our training strategy and data ratio. Key experiments included a $\mu$P hyperparameter search and data ratio optimization, which confirmed the strategy's suitability for \llm.

\subsection{$\mu$P Hyperparameter Search}

\begin{figure}[htbp]
    \centering
    \includegraphics[width=0.8\linewidth]{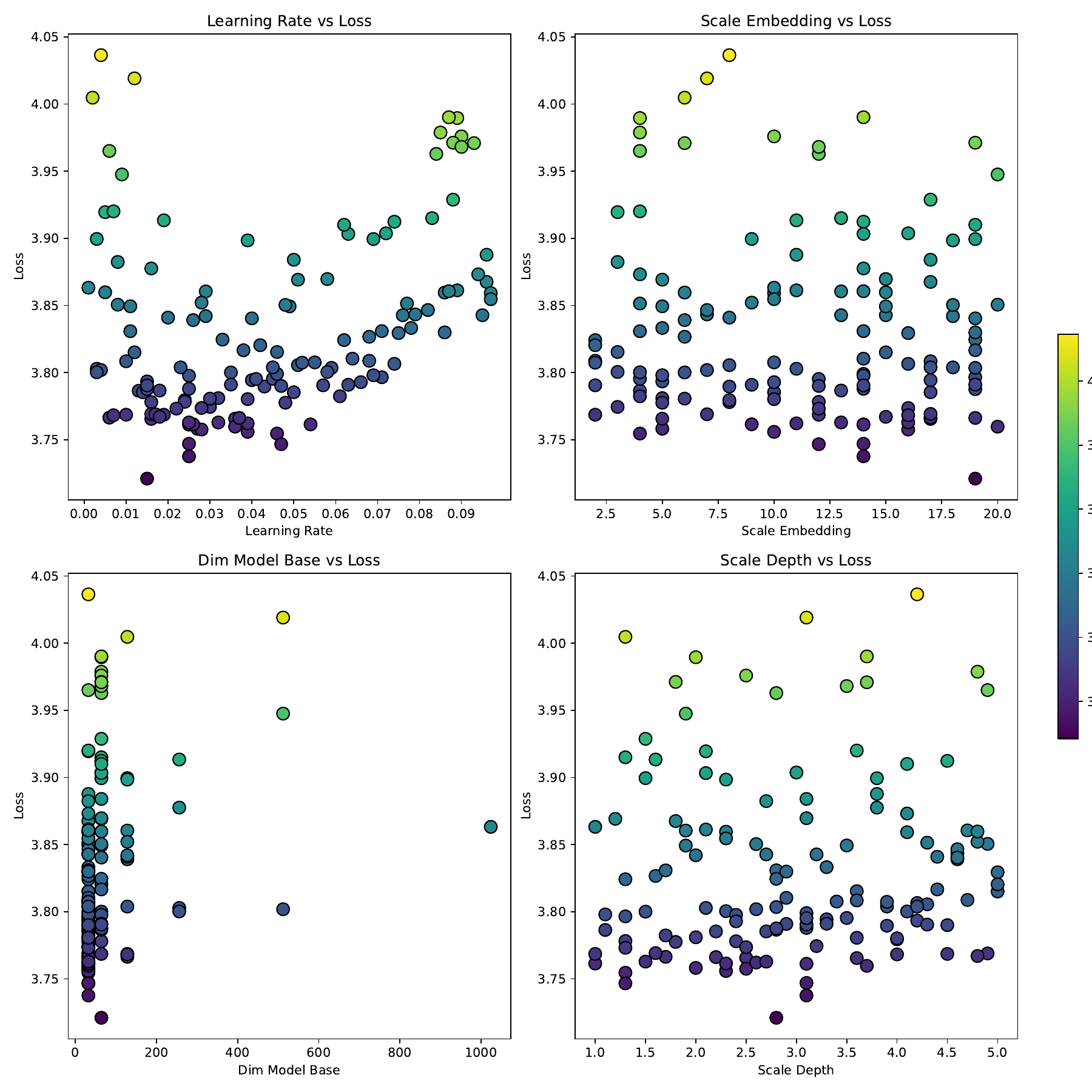}
    \caption{Grid search over the $\mu$P parameterization spaces.}
    \label{fig:mupsearch_app}
\end{figure}

We observed that $\mu$P hyperparameters remained stable across model scales. Using Bayesian optimization, we optimized four key hyperparameters: $scale\_emb$, $dim\_model\_base$, $scale\_depth$, and $learning\_rate$ on the nano model with the C4 dataset. The search explored 300 configurations, compared to 570,000 in a grid search. Results showed:


\begin{itemize}
    \item \textbf{Optimal Hyperparameters:} $learning\_rate$ between 0.01 and 0.02, and $dim\_model\_base$ below 256. 
    \item \textbf{Loss Patterns:} Loss below 4.1 concentrated around specific hyperparameters, indicating stable performance (Figure~\ref{fig:mupsearch_app}).
\end{itemize}


\begin{table}[ht]
  \centering
  \setlength{\tabcolsep}{12pt} 
  \renewcommand{\arraystretch}{0.6} 
  \begin{tabular}{@{}lccc@{}}
    \toprule
    \textbf{Hyperparameter} & \textbf{Range} & \textbf{Options} & \textbf{Step Size} \\
    \midrule
    \texttt{scale\_emb} & [2, 20] & - & 1 \\
    \texttt{dim\_model\_base} & - & \{32, 64, 128, 256, 512, 1024\} & - \\
    \texttt{scale\_depth} & [1, 5] & - & 0.1 \\
    \texttt{learning\_rate} & [0.001, 0.1] & - & 0.001 \\
    \bottomrule
  \end{tabular}
  \caption{Hyperparameter search ranges for nano model.}
  \label{tab:hyperparameter_ranges}
\end{table}

\begin{table}[h]
\centering
\begin{tabular}{l|p{8cm}}
\toprule
\textbf{Name}                       & \textbf{Specific Operation}                                                                                                                 \\ \midrule
Embedding Output Scaling            & Multiply the output of the embedding by $scale\_{emb}$                                                                                                                                                     \\ \hline
Residual Connection Scaling         & Scale the output tensor of a block before adding to each residual connection in each layer by $scale\_{depth}/\sqrt{\text{num\_layers}}$ 
\\ \hline
Initialization of Tensors           & Set the initialization standard deviation of each two-dimensional tensor parameter to $init\_std/\sqrt{d_m/d_{base}}$, and set other parameters' initialization to 0.1                  \\ \hline
Learning Rate Scaling of Tensors    & Adjust the learning rate of each two-dimensional tensor parameter to $1/({d_m/d_{base}}) $ times the learning rate of other parts (or the overall learning rate)                    \\ \hline
LM Head Scaling                    & Adjust the output logits to $1/(d_m/d_{base})$ times the original value                                                                                                           \\ \bottomrule
\end{tabular}
\caption{List of operations used when applying tensor program techniques. }
\label{tab:mup}
\end{table}


\end{document}